\documentclass[double]{article}
\usepackage{times}         
\usepackage{graphicx}       
\usepackage{epsfig}         
\usepackage{tikz}          
  \usetikzlibrary{shapes.geometric, arrows}
\usepackage{amsmath, amssymb, amsthm, amsfonts}
\usepackage{array}         
\usepackage{algorithm}
\usepackage{algorithmic}
\usepackage{booktabs, array}
\usepackage{longtable, multirow}
\usepackage{siunitx}          
\usepackage{booktabs}       
\usepackage{tabularx}       
\usepackage{multirow}      
\usepackage{longtable}     
\usepackage{makecell}     
\usepackage{adjustbox}      
\usepackage{float}        
\usepackage{subcaption}    
\usepackage{wrapfig}        
\usepackage{verbatim}       
\usepackage{listings}      
\usepackage{dirtytalk}     
\usepackage{xcolor}         
\usepackage{lineno}         
\usepackage{setspace}       
\usepackage{comment}       
\usepackage{srcltx}         
\usepackage{hyperref}       
  \hypersetup{
    colorlinks=true,
    linkcolor=blue,
    citecolor=teal,
    urlcolor=magenta
  }
\usepackage[
  backend=biber,
  style=numeric,
  sorting=none,
  maxnames=9,
  minnames=1,
  giveninits=true
]{biblatex}
\usepackage{authblk}      
\newcolumntype{L}{>{\raggedright\arraybackslash}p{3.4cm}} 
\newcolumntype{C}{>{\centering\arraybackslash}p{2.9cm}}

\begin{document}

\title{\Large Knowledge Integration for Physics-informed Symbolic Regression Using Pre-trained Large Language Models}

\author{Bilge Taskin$^{1}$, Wenxiong Xie$^{1}$, Teddy Lazebnik$^{2,1,*}$\\ \(^1\) Department of Computing, Jonkoping University, Jonkoping, Sweden \\ \(^2\) Department of Information Science,  University of Haifa, Haifa, Israel \\ \(^*\) Corresponding author: lazebnik.teddy@gmail.com \\ }

\date{ }

\maketitle 

\begin{abstract}
Symbolic regression (SR) has emerged as a powerful tool for automated scientific discovery, enabling the derivation of governing equations from experimental data. A growing body of work illustrates the promise of integrating domain knowledge into the SR to improve the discovered equation's generality and usefulness. Physics-informed SR (PiSR) addresses this by incorporating domain knowledge, but current methods often require specialized formulations and manual feature engineering, limiting their adaptability only to domain experts. In this study, we leverage pre-trained Large Language Models (LLMs) to facilitate knowledge integration in PiSR. By harnessing the contextual understanding of LLMs trained on vast scientific literature, we aim to automate the incorporation of domain knowledge, reducing the need for manual intervention and making the process more accessible to a broader range of scientific problems. Namely, the LLM is integrated into the SR's loss function, adding a term of the LLM's evaluation of the SR's produced equation. We extensively evaluate our method using three SR algorithms (DEAP, gplearn, and PySR) and three pre-trained LLMs (Falcon, Mistral, and LLama 2) across three physical dynamics (dropping ball, simple harmonic motion, and electromagnetic wave). The results demonstrate that LLM integration consistently improves the reconstruction of physical dynamics from data, enhancing the robustness of SR models to noise and complexity. We further explore the impact of prompt engineering, finding that more informative prompts significantly improve performance. \noindent \\ \\ 

\end{abstract}

\maketitle \thispagestyle{empty}
\pagestyle{myheadings} \markboth{Draft:  \today}{Draft:  \today}
\setcounter{page}{1}

\onehalfspacing

\section{Introduction}
When Isaac Newton formulated his laws of motion and universal gravitation in the 17th century, he demonstrated the extraordinary power of mathematical expressions to capture physical phenomena \cite{intro_1}. Newton's work represented a watershed moment, showing how complex observations could be distilled into elegant mathematical equations that not only described but predicted natural phenomena with remarkable accuracy \cite{intro_2,intro_3,intro_4}. In a more general sense, scientific discovery has long been driven by the interplay between empirical observation and theoretical modeling \cite{intro_5}. This fundamental relationship has guided our understanding of the natural world since the inception of modern science \cite{intro_6}. 

Throughout the subsequent centuries, this approach of expressing scientific knowledge through mathematical relationships has remained central to physics and other natural sciences \cite{intro_7,intro_8,intro_9}. From Maxwell's equations unifying electricity and magnetism \cite{intro_10} to Einstein's theories of relativity redefining our understanding of space and time \cite{intro_11}, mathematical formulations have consistently provided the language through which we articulate physical laws \cite{intro_12}. These symbolic expressions serve not merely as descriptive tools but as engines of discovery \cite{intro_13}, enabling scientists to derive new insights and make predictions beyond what has been directly observed.

The process of deriving mathematical expressions from experimental data, formally known as symbolic regression (SR), has traditionally been a labor-intensive endeavor requiring deep domain expertise and mathematical intuition \cite{intro_14,intro_15,intro_16}. Scientists have historically relied on a combination of theoretical understanding, creative insight, and iterative refinement to formulate equations that accurately capture observed phenomena \cite{intro_17,intro_18}. This approach, while being the central force in modern science, is constrained by human cognitive limitations and prior theoretical frameworks.

In recent years, the emergence of machine learning (ML) and artificial intelligence (AI) has significantly accelerated the process of scientific knowledge extraction, particularly in domains that require complex pattern recognition and hypothesis generation \cite{intro_19,intro_20,intro_21,intro_22}. These advantages occurred hand in hand with the exponential increase in available scientific data across all fields, including medicine \cite{intro_23}, economics \cite{intro_24}, and physics \cite{intro_25}, to name a few. Despite their established benefits \cite{intro_27,intro_28}, current AI, in general, and ML models, in particular, are not explainable and lack the reproducibility and generalizability associated with symbolic mathematical expressions \cite{intro_26,intro_29}. 

To address this gap, two main branches have emerged over the last several years, one aims to make AI models more explainable \cite{intro_30,intro_31,intro_32}, while the other tries to automate the SR process using computational methods \cite{intro_33}. Unlike traditional regression techniques, SR does not assume a predefined model structure but instead searches for the most appropriate mathematical representation using optimization techniques \cite{intro_34}. Recently, automatic scientific discovery has benefited from the integration of AI and data-driven methods, enabling researchers to uncover new scientific laws and relationships with less human intervention \cite{intro_35,intro_36}. In physics, this process is particularly relevant as it allows for the derivation of governing equations from raw experimental or simulated data, offering a data-centric approach to understanding physical systems \cite{intro_37}. Indeed, SR has proven to be a valuable tool in this endeavor, capable of rediscovering known physical laws, such as Newton’s laws of motion and Maxwell’s equations from observational  data~\cite{Schmidt2009,Stalzer2019}. However, despite its potential, SR faces significant challenges, particularly in ensuring that the discovered expressions remain interpretable and physically meaningful rather than simply fitting a given experimental data \cite{cranmer2020discovering, udrescu2020ai, brunton2016discovering}.

To address these challenges, recent studies have explored physics-informed SR (PiSR), which incorporates domain knowledge to guide the discovery process \cite{scimed}. These approaches leverage physical constraints, symmetries, and conservation laws to improve the quality of the inferred models. By embedding these principles into the learning process, PiSR enhances the plausibility and generalizability of discovered equations \cite{intro_38,intro_39}. However, the integration of domain knowledge into PiSR remains highly technical, often requiring specialized formulations, manual feature engineering, and handcrafted constraints, making it difficult to generalize across different physical domains \cite{scimed}.

In this study, we propose a novel approach by leveraging pre-trained Large Language Models (LLMs), a family of deep learning models trained on vast textual and scientific datasets to develop a broad understanding of language (enabling them to generate, analyze, and integrate domain-specific knowledge in various tasks), to facilitate knowledge integration in PiSR. LLMs, trained on vast corpora of scientific literature and equations, have demonstrated remarkable capabilities in understanding and generating scientific content \cite{intro_40}. By harnessing the contextual understanding of LLMs, we aim to automate the incorporation of domain knowledge in SR, reducing the need for manual intervention and making the process more accessible to a broader range of scientific problems. 

The rest of the manuscript is organized as follows. Section \ref{sec:rw} presents the current state-of-the-art in SR and LLMs, as well as previous attempts to integrate the two. Next, section \ref{sec:method} formally introduces the proposed LLM and SR integration mechanism. Following that, section \ref{sec:experiments} describes the experimental setup used to evaluate the proposed method. Afterward, section \ref{sec:results} outlines the obtained results. Finally, Section \ref{sec:discussion} discuss the results in an applied context and suggests possible future work. 
    
\section{Related Work}
\label{sec:rw}
This section provides an overview of research related to automatic scientific discovery, SR, and LLMs. We initially discussed automatic scientific discovery. Next, we examine the current state-of-the-art SR with a particular focus on PiSR. We then review recent advancements in large language models. Finally, we explore the integration of LLMs and SR models.

\subsection{Automatic scientific discovery}
Automated scientific discovery (ASD) has gained increasing attention and played an important role in the field of intelligent science \cite{asd_8,asd_9,asd_10}. ASD differs from traditional manual scientific discovery in that it aims to obtain new scientific knowledge with minimal human intervention \cite{asd_11}. The development of ASD is closely linked to advances in automation, data processing, and AI, which have contributed to a paradigm shift in the way scientific research is conducted \cite{asd_12,asd_13,asd_14}. 

The field of ASD has inspired the imagination of scholars for several decades, with initial results as early as the late 1970s \cite{asd_20}. Specifically, Pat Langley developed the first data-based equation finding system \cite{asd_21}, which found equations by monitoring correlations between pairs of variables and generating new terms \cite{asd_22}. A decade later, the Lagrange system is introduced, which can process observed data and learn dynamic system models in the form of ordinary differential equations \cite{asd_23}. Around the same time, Koza introduced the concept of SR in automatic scientific discovery \cite{asd_24} alongside Petrovski and Džeroski, who used SR methods to discover ordinary differential equations \cite{asd_25}. A decade and a half ago, \cite{asd_1} developed a robot scientist which was able to discover six genes encoding orphan enzymes in yeast \cite{asd_2}, marking a turning point in ASD, as this solution is not only able to formulate hypotheses without human intervention, but also to test and iterate discoveries through automated experiments \cite{asd_1,asd_15}. Recently, advanced computational methods such as SR and LLM emerging as the next phase of ASD \cite{asd_6}. 

\subsection{Symbolic regression}
SR is a modeling technique for discovering mathematical expressions that represent the relationship between a set of input variables and an output variable \cite{sr_12,sr_13,sr_14}. SR does not require a predefined model structure, unlike other regression models, such as linear regression, as it searches through a large number of potential mathematical expressions to identify the function that best fits the given data \cite{sr_15}. This approach is particularly useful in cases where the underlying functional form is unknown or complex \cite{sr_1,sr_2,sr_3}. 

Roughly speaking, one can categorize the SR methods into four main groups based on the underlying computational techniques: brute force, sparse regression, deep learning, and genetic algorithms \cite{sr_16}. The brute-force-based SR methods can theoretically solves any SR tasks. Nevertheless, practically, their computational cost is unrealistic, and they are extremely prone to overfitting \cite{sr_23}. The method tests all possible equations to find the one that performs optimally, differing in the way they cover the search space \cite{sr_22}. The sparse regression SR method, which identifies parsimonious models with the help of optimization that promotes sparsity, significantly reduces the search space. Among them, SINDy is designed for scientific use cases, which employs the Lasso linear model for sparse identification of nonlinear dynamical systems behind time series data \cite{sr_20}. SINDy iterates between partial least squares fitting and thresholding (promoting sparsity) steps \cite{sr_21}. Deep learning SR methods excel in handling noisy data due to the high outlier resistance of neural networks, but are limited in their generability \cite{LaCava2021SRBench}. These methods use neural networks to produce analytical equations. For instance, \cite{sr_17} proposed a Deep Symbolic Regression (DSR) system for general SR tasks, utilizing reinforcement learning to train a generative recurrent neural network (RNN) model of symbolic expressions and employs a variant of the Monte Carlo policy gradient technique called \say{risk-seeking policy gradient} to adapt the generative model to the exact formulation. Lastly, genetic-algorithm-based SR methods treat mathematical expressions as individuals (or genes) in a population \cite{sr_16}. These individuals evolve over time through mechanisms such as selection, crossover, and mutation, gradually refining the equations that can better fit the data. This evolutionary approach allows SR to discover interpretable models in a flexible way, without being constrained by predefined assumptions \cite{sr_4,sr_5}. For example, the gplearn Python library implements a genetic algorithm for SR, showing a promising ability to re-discover known physical equations \cite{sr_19}. It first constructs a population of stochastic formulas representing the relationship between known independent variables (features) and dependent variables (objectives), presented in a tree structure. These subtrees are then replaced and restructured in a stochastic optimization process that computes fitness by executing the tree and evaluating its output, using a stochastic strategy of survival of the fittest.

In recent years, many SR algorithms have been proposed~\cite{tohme2022gsr,landajuela2022udsr,jiang2024vsrdpg}.From these, several SR algorithms gain much popularity and provide a simple programing interface to use, including the DEAP (Distributed Evolutionary Algorithm in Python) \cite{sr_6,sr_7}, gplearn \cite{sr_8,sr_9}, and PySR \cite{morales2024}. All three algorithms belong to the genetic-algorithm-based SR family of methods. 

\subsection{Physics-informed symbolic regression}
SR is widely adopted for physical problems, which often face a multivariate and noisy experimental data from nonlinear systems and require analytical equations to discover general laws of nature \cite{asd_4,asd_5,asd_7}. In this context, simply using SR methods provides limited results as the SR needs to make sure the prediction equation follows scale consistency \cite{sr_16} along with other properties. To this end, Physics-Informed Symbolic Regression (PiSR) has emerged, aiming to close this gap \cite{pisr_5,asd_7,sr_16}. PiSR is designed to guide the search process of symbolic expressions by incorporating physics principles such as conservation laws, symmetry, and sparsity assumptions, to name a few \cite{asd_7}. 

For example, the AI Feynman SR algorithm combines physics-inspired techniques with machine learning and deep learning methods to improve the accuracy of physics discoveries, showing promising results across multiples sub-field of physics \cite{udrescu2020ai}. Similarly, \cite{pisr_1} proposed Equation Learner (EQL) which is based on a multilayer feedforward neural network designed to learn the dynamical equations of a physical system and to be able to extrapolate to unseen domains. It allows efficient training based on gradients, and its architecture consists of linear mappings, nonlinear transformations consisting of specific unitary and binary (multiplicative) units, and a linear readout layer. \cite{pisr_2} proposed the Scalable Pruning for Rapid Identification of Null vecTors (SPRINT) algorithm, which is a model-independent sparse regression algorithm that extends the iterative singular value decomposition (SVD) based exhaustive search algorithm. 

\subsection{Large language models and symbolic regression}
Large language models (LLMs) have recently gained significant traction, often surpassing traditional approaches in multiple tasks in information retrieval, information representation, and natural language processing \cite{hou2024large, wu2023survey, zhao2024recommender,llm_8,llm_9,llm_10}. Research~\cite{kim2024large, liu2024once} combining LLMs with established techniques such as collaborative filtering and content-based methods has yielded highly accurate recommendations. Furthermore, LLMs are proving valuable in improving specific recommendation challenges, including sequential recommendations~\cite{harte2023leveraging, li2024large}, long-tail recommendations~\cite{wu2024coral}, and cold-start scenarios~\cite{sanner2023large}.

Building on their success in several domains \cite{llm_good_1,llm_good_2,llm_good_3,llm_good_4}, LLMs are also being explored for their potential to enhance knowledge integration in various downstream tasks \cite{wei2021pretrained}. Their ability to understand and generate human-like text, coupled with their vast pre-trained knowledge, makes them promising candidates for incorporating external knowledge into models for tasks like physical symbolic regression \cite{ils_2}. This involves leveraging LLMs to not only understand the underlying physical principles and relationships expressed in text but also to guide the search for symbolic equations that accurately describe observed data \cite{sharlin2024context}. By integrating knowledge from scientific literature, textbooks, or even expert explanations, LLMs could potentially help overcome limitations of purely data-driven symbolic regression methods, leading to more interpretable and physically plausible models.

\cite{ils_1} show that while existing SR methods mostly rely on techniques such as genetic algorithms and Markov chain Monte Carlo sampling, they lack understanding of the problem's domain. Even recent Transformer-based SR methods are limited in the context they are used for since they learn pattern matching between the dataset and the mathematical expression they have been provided with, limiting their view of similar problems that can be used. In order to tackle this challenge, the authors used the GPT-4 LLM (and GPT-4o) model to propose expressions based on the data, which were then optimized and evaluated using external Python tools, and then the results were fed back to the LLM to propose better expressions. Specifically, they added two hints to the LLM: One is the initial prompt, which is to input data for the model to suggest expressions; the second hint is the iterative hint, which is to combine the data and feedback for the model to optimize the expression.  \cite{ils_2} investigated the integration of LLMs into the SR process, utilizing contextual learning (ICL) capabilities. The authors proposed the ICSR method, which is inspired by the Optimization by Prompting (OPRO) framework. That is, by providing LLMs with previously tested equations and fitness scores, they are made to generate better candidate equations and iterate until convergence or reach the computational budget. Using LLaMA3 8B as the underlying LLM, the ICSR method performed comparably well in all current popular SR benchmarks and generates expressions with low average complexity. \cite{ils_3} focused on how to simplify the complexity of finding expressions reflecting the relationships between variables from observed data. The authors automatically generate expressions by describing requirements in natural language to an LLM. In particular, the authors proposed the MLLM-SR method, which uses LLMs to generate, run SR methods, and repeat until automatically-generated conditions are met. 

In all these studies, the LLM is combined in a pipeline before or after the SR's equation prediction. In this study, we aim to integrate LLM into the SR's search process itself. 

\section{LLM-based Knowledge Integration to SR}
\label{sec:method}
In this study, we propose the integration of SR with LLMs to address to provide a physically-informed context and straightforward knowledge integration of the expert user to the SR's search process. We hypothesis that thanks to the information LLMs contain from a wide range of scientific texts they trained on, they can evaluate the equations produced by SR in terms of dimensional consistency, scientific validity, and domain knowledge. Using this capability, LLM can provide a guiding term to the SR's loss function. 

\begin{figure}
    \centering
    \includegraphics[width=0.99\linewidth]{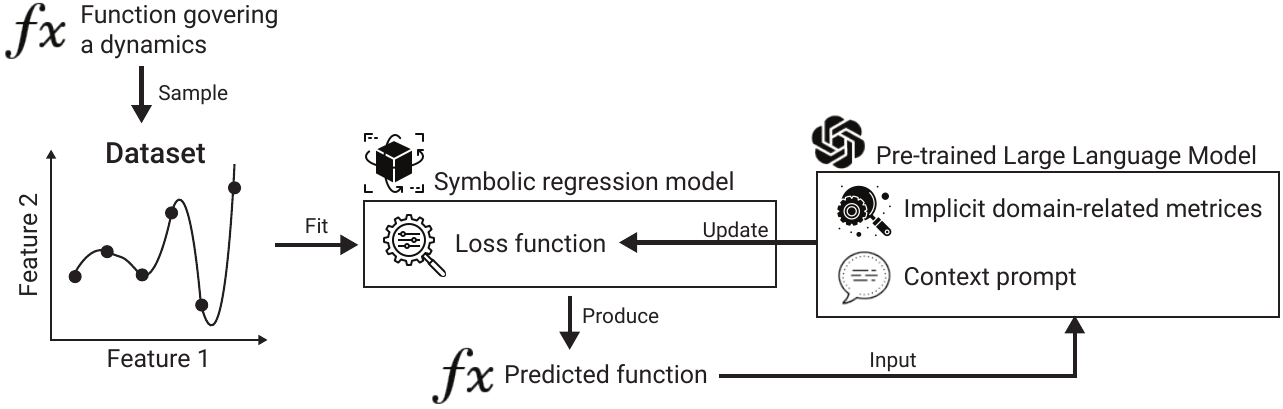}
    \caption{A schematic view of the Large Language Model-based Knowledge Integration into the Symbolic Regression's searching (and optimization) process.}
    \label{fig:scheme}
\end{figure}

To this end, let us assume an optimization system with two components, an SR and an LLM, denoted \(S\) and \(M\), respectively. Given a dataset \(D \subset \mathbb{R}^{n \times m}\) with \(n\) samples and \(m\) features, we wish to find an analytical equation that fits the data while also aligning with pre-defined list of physical constraints, represented as a metric functions accepting an equation as an expression tree and aware of the LLM's implicit knowledge. To this end, the SR component receives the dataset and a loss function \(L\) that expects a dataset, an analytical function (represented as an expression tree; \cite{koza1992genetic, schmidt2009distilling}) and returns a value in \(\mathbb{R}^+\) which indicates how well the analytical function fits and explains the provided dataset. Formally, we propose the following loss function:
\begin{equation}
L = w_1  e + w_2 s + w_3 c,
\label{eq:loss}
\end{equation}
where \(w_1, w_2, w_3 \in [0, 1] \wedge w_1 + w_2 + w_3 = 1\), \(e\) is the mean square error \cite{schmidt2009distilling} between the prediction of the provided analytical function and the provided dataset, \(s\) is the size of the analytical equation in terms of nodes in the expression tree, and \(c\) is the LLM-based score of how well the analytical equation fulfills a pre-defined list of physical constraints. Fig. \ref{fig:scheme} shows a schematic view of the LLM-based knowledge integration into SR.

Importantly, in order to ensure the reliability of the LLM-based score, which is not necessarily can be a floating point value but rather any string, we carefully design the LLM's prompt. To be exact, we used the following prompt:

\lstset{
    breaklines=true,
    breakatwhitespace=true,
    basicstyle=\ttfamily\small,
    backgroundcolor=\color{gray!10},
    frame=single
}

\begin{lstlisting}
PROMPT = f"""
### ROLE
You are an expert *scientific-reasoning* assistant.
Return **ONLY** a Python-style list:
[dim_corr, simp, sim, "feedback"].

### METRICS
dim_corr . 0 (wrong) -> 1 (perfect)
simp     . 0 (complex) -> 1 (simple)
sim      . 0 (unrealistic) -> 1 (realistic)


### FEW-SHOT EXAMPLES
#1  Equation:  x = v0 * t + 0.5 * g * t^2  
    Output:    [0.95, 0.80, 0.92, "Classic kinematics"]
#2  Equation:  E = m + c  
    Output:    [0.05, 0.70, 0.15, "Units mismatch"]
#3  Equation:  y = sin(sin(x))  
    Output:    [0.90, 0.10, 0.40, "Needless nesting"]

### TASK
Equation to evaluate:
{equation_str}

{f"Context:\n{context}" if context else ""}

Think step-by-step silently, then output the list only.
"""
\end{lstlisting}

To ensure that the prompt would work consistently across different models, we adopted a few key design principles. First, we set the temperature to zero to guarantee deterministic outputs - ensuring that models like Mistral, Llama, and Falcon would always produce the same response for the same input. We introduced a ``\#\#\# ROLE'' section to immediately frame the assistant’s focus on scientific reasoning, helping the model interpret the task correctly from the start. In the ``\#\#\# METRICS'' section, we clearly defined three core evaluation criteria --- dimensional consistency, simplicity, and physical realism --- using well-spaced numerical scales so that all models could interpret them in the same way. We included a few carefully chosen examples (a classic kinematics equation, a dimensional mismatch, and an unnecessarily nested trigonometric form) to anchor the scoring logic. Finally, we enforced a strict Python-style list format for the outputs and emphasized the ``ONLY'' instruction to minimize the chance of extra text generation. With these measures in place, the [dim\_corr, simp, sim, feedback] lists returned by Mistral, Llama, and Falcon were not only structurally consistent but also reliably comparable.

To be exact, upon receiving the prompt, each LLM returns a concise, three-element list, including a score about physical dimensionality (\(c_1\)) consistency, syntactic and structural simplicity (\(c_2\)), and physical realism (\(c_3\)). The LLM's overall score is computed as follows:
\begin{equation*}
    c := 1 - (c_1 + c_2 + c_3)/3.
\end{equation*}

\section{Experiments}
\label{sec:experiments}
\begin{figure}[!ht]
    \centering
    \includegraphics[width=0.99\linewidth]{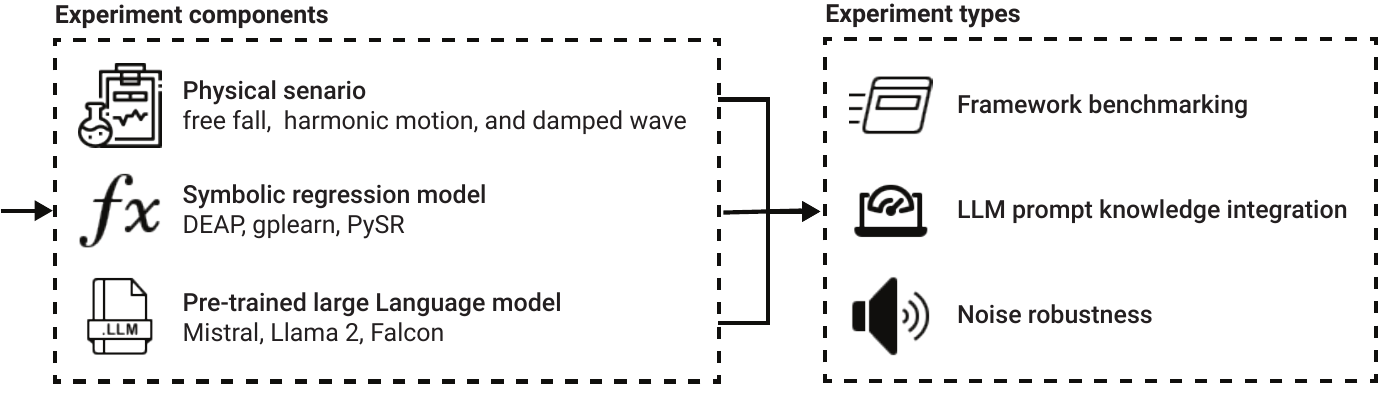}
    \caption{A schematic view of the experimental setup.}
    \label{fig:experiment}
\end{figure}

In order to evaluate the proposed method, we designed an experiment across three \textit{in silico} physical scenarios (free fall, simple harmonic motion, damped waves), using three symbolic regression implementations (PySR, DEAP-GP, gplearn) and three language models (Mistral 7B, Llama 2 7B, Falcon 7B), evaluating the robustness of the proposed method over all three axes. For each combination of these components, we conducted three experiments - benchmarking the performance, comparision between different knowledge integration provided to the LLM's prompt, and noise robustness. For all of these experiments we recorded four main metrics: mean absolute error (MAE), mean square error (MSE), coefficient of determination (\( R^2 \)), and the expression tree score. Fig. \ref{fig:experiment} presents a schematic view of the experimental setup.

\subsection{Experiment components}
\subsubsection{In Silico Physical Experiments}
\label{sec:data}
As a representative set of physical problems, we adopted classical Newtonian movement cases with an increasing level of reconstruction complexity, as reflected by the number of nodes in their expression trees. All equations show non-linear dynamics, ensuring a real-world level of complexity. Namely, the three physical cases are: free fall of a ball, simple harmonic motion, and damped wave—were.  First, consider a freely falling object near Earth's surface under constant gravity. 
When dropped from rest at height~$h\,[\mathrm{m}]$, its speed follows
\begin{equation}
v = \sqrt{2 g h},
\end{equation}
where $g = 9.81\,\mathrm{m/s^{2}}$. 
Next, the simple harmonic motion takes the following form:
\begin{equation}
    x(t) = A \cos\Bigl(\tfrac{k}{m}t + \varphi\Bigr),
\end{equation}
with $A$ as amplitude, $k$ the spring constant, $m$ the mass, and $\varphi$ the phase offset \cite{goldstein2001}. 
Finally, a damped plane-wave propagation at fixed position $x$. The field obeys the following equation:
\begin{equation}
E(t) = E_0 \, e^{-\alpha t/2}\,\cos\bigl(kx - \omega t\bigr),    
\end{equation}
where $\alpha$ is the damping coefficient \cite{jackson1999}. 
From now on, we would refer to these equations as the \say{ground truth} (GT) equations.

In order to generate synthetic data for each of the \textit{in silico} physical
scenarios, we implemented the following procedure. Initially, we conducted
\(N = 500\) experiments, uniformly sampling five independent parameters for
their initial condition, including:
(1) mass, randomly selected from \(0.1~\mathrm{kg}\) to \(10~\mathrm{kg}\);
(2) a characteristic length (e.g.\ drop height, spring amplitude, or radius)
between \(0.01~\mathrm{m}\) and \(0.5~\mathrm{m}\);
(3) initial height/displacement, varying from \(1~\mathrm{m}\) to
\(100~\mathrm{m}\);
(4) a linear drag/damping coefficient from \(0\) to \(1~\mathrm{kg\,s^{-1}}\); and
(5) time, sampled randomly between zero and the physically realistic maximum
of \(\sqrt{2h/g}\), where \(g = 9.81\,\mathrm{m\,s^{-2}}\).
Each experiment is run until reaching equilibrium or 1000 steps in time.
This approach provides a broad and realistic representation of
conditions~\cite{Keijzer2003,Uy2011}. 

With parameters determined, we proceeded to generate ground truth data using analytical equations corresponding to each physical scenario. These calculations provided precise, idealized responses against which the model's performance could be evaluated. For example, free fall velocity was calculated as  \( v = \sqrt{2gh} \) to validate the model's handling of square-root dependencies. Harmonic oscillator displacement was computed using  \( x = A \cos(\sqrt{k/m}\,t + \phi) \) to assess the model's capture of trigonometric functions. Damped wave magnitude was calculated according to \( E = E_0 e^{-\alpha t/2}\cos(kx - \omega t) \) to test the model's ability to model exponential decay and oscillation. The results of these calculations served as our ground truth \cite{Schmidt2009}.

Recognizing the inevitability of measurement imperfections in practical settings, we added zero-mean Gaussian noise to our ideal signals. The noise level was set to 1\% of each signal's variability, resulting in an average signal-to-noise ratio (SNR) of approximately 40 dB, a value representative of typical laboratory instrumentation. This process yielded noisy observations, denoted by \(y\), which incorporate a realistic level of measurement uncertainty \cite{Vladislavleva2009,Reinbold2021}. With the five independent parameters defined, we constructed a predictor matrix \(X\). The corresponding noisy signals were assigned as the target vector \(y\). The data were retained in their original SI units, relying on the regression models to manage any required scaling, normalization, or transformations internally \cite{Keijzer2003}.

\subsubsection{Large language models}
\label{subsubsec:llm}
We used three open-source and locally-depolyable LLMs - Mistral 7B, Llama 2 7B, and Falcon 7B. We picked these models due to their relative small size, which allows to run them locally on standard computer system on the one hand and their reasoning and robust training process, on the other hand. While other LLMs such as Gemini 2 and GPT-o1 that requires API (Application Programming Interface) calls may produce better results, we decided to use local LLMs as they both cheaper and ensures sensitive data is not shared with 3rd party in practical usage \cite{shojaee2025llmsr, romera2023funsearch, merler2024icsr}. Importantly, we set the sampling temperature to zero to guarantee deterministic, repeatable outputs and to ensure that differences in plausibility scores arise solely from model knowledge rather than random sampling noise \cite{rae2021scaling,openai2023gpt4,chen2023frugal}. 

\subsubsection{Symbolic regression models}
\label{subsubsec:sr-models}
We used three popular SR models: DEAP, gplearn, and PySR. A uniform population size of 100 and 50 generation (iterations) are used for all three SR models to balance between a robust search process and computational resources. Specifically, for PySR, we used the Huber loss function and enabled a complexity penalty that softly discourages extra operators, yielding more concise formulas. For DEAP, we controlled genetic variation with a crossover probability of 0.6, a mutation probability of 0.05, tournament selection size of 3, and a maximum tree depth of 8 to prevent bloat. Finally, for gplearn, we applied a substantial complexity penalty on operator counts so that only genuinely informative structures preserved. Across all three SR models, we utilized the following operator set: $+$, $-$, $\times$, $\div$, $\exp$, $\log$, $\sin$, and $\cos$. Table~\ref{tab:hyperparams}) summarizes the used hyperparameters and their values for these models.

\begin{table}[H]
  \centering
  \caption{Hyperparameter settings for symbolic regression tools.}
  \label{tab:hyperparams}

  \begin{tabular}{@{}L L c c c@{}}
    \toprule
    \textbf{Category} & \textbf{Hyperparameter} & \textbf{PySR} & \textbf{DEAP--GP} & \textbf{gplearn} \\ 
    \midrule

    \multirow{2}{*}{General}
      & Population size          & 100 & 100 & 100 \\
      & Iterations / Generations & 50  & 50  & 50 \\
    \midrule

    \multirow{3}{*}{Genetic Settings}
      & Crossover probability & --- & 0.05 & --- \\
      & Mutation probability  & --- & 0.01 & --- \\
      & Tournament size       & --- & 3    & --- \\
    \midrule

    \multirow{2}{*}{Tree Settings}
      & Tree–height limit & --- & 8 & --- \\
      & Initial tree depth & --- & 1 & --- \\
    \midrule

    \multirow{2}{*}{Operators}
      & Binary operators &
        \makecell{+, –, *, /,\\$\wedge$} &
        \makecell{+, –, *, /,\\protected\_div} &
        \makecell{add, sub,\\mul, div} \\
    \cmidrule(lr){2-5}
      & Unary operators &
        \makecell{neg, exp, log,\\sin, cos} &
        \makecell{neg, log,\\sin, cos} &
        \makecell{neg, exp, log,\\sin, cos} \\
    \bottomrule
  \end{tabular}
\end{table}

Notably, to prevent overfitting and improve computational efficiency, we used an early-stopping rule. Optimization stopped if the best composite loss improved by less than 0.1\% over three generations \cite{Yuan2024EarlyStopping}.

\subsection{Experiment types}
\subsubsection{Benchmarking}
For the benchmarking of the proposed LLM-integrated SR framework, we computed all 27 (3 LLMs, 3 SR, and 3 physical scenarios) configuration and compared them to the baseline configuration of each SR model in which no LLM model is used. 

\subsubsection{LLM prompt knowledge integration experiment}
\label{subsubsec:llm-prompt-sensitivity}
Due to the fact that the exact prompt used to query the LLM results can highly different responses, and therefore results, we explored the influence of different types of knowledge in prompts on the LLM's influence on the SR's produced equation. To this end, we consider eight configurations. Formally, prompt A provided no additional information, acting as a baseline. Prompt B included data-column descriptions only, testing if this reminder of physical quantities improved dimensional reasoning. Prompt C added a concise, natural-language \say{Experiment description}, exploring whether a broader physical context improved plausibility judgments \cite{zhu2023promptbench}. Prompt D appended the GT equation, investigating whether this reiteration of the ground truth induced alignment bias \cite{cao2024worst}. In addition to the single-cue prompts, we explored combinations to investigate synergistic or competing effects. Prompt E merged B and C to assess whether pairing raw data details with contextual framing yields improved performance. Prompt F merged B and D to test whether coupling data reminders with the true formula enhances dimensional correction, potentially at the risk of over-guiding the LLM \cite{sclar2023quantifying}. Prompt G merged C and D to evaluate if contextual description combined with ground-truth disclosure facilitates reconciling intuition with explicit answer keys \cite{raj2023semantic}. Finally, Prompt H fused B, C, and D, creating a \say{kitchen-sink} template to determine whether information overload hinders performance or if redundancy reinforces consistent scoring \cite{zhuo2024prosa}. The prompt variant (A–H) used for the three physical scenarios is provided as supplementary material.

\subsubsection{Noise robustness experiment}
\label{subsubsec:data-sensitivity}
To assess how our LLM-integrated SR model holds up when measurements get noisy, we followed the experiment design proposed by \cite{keren2023computational}. Namely, we first followed the regular data generation procedure (see Section \ref{sec:data}) but without adding noise. We then added zero-mean Gaussian noise at five levels— 1 \%, 2\%, 3\%, 4\%, and 5\% of each signal’s standard deviation—applying it separately to the input features and to the target variable. 

\subsection{Expression tree score metric}
Evaluating equations only with numerical error values (e.g., mean square error or coefficient of determination) answers merely the question of \say{how well they fit the data}. However, our goal is to determine how close the discovered formulas are to the data numerically and how structurally similar they are to the real underlying formula. For this purpose, we use the \textit{expression tree distance} method to compare each equation we obtain with the GT equations \cite{matsubara2024rethinking, Bertschinger2023Metric}. 

We compute the expression tree distance by first converting equations to expression trees \cite{LaCava2021SRBench}, where operations are arranged hierarchically (e.g., "+", "-", "*"). Internal nodes represent operators with left/right subtrees, while leaf nodes contain numbers or variables. The distance is measured by simultaneously comparing the trees from the root. Identical leaf node contents yield a distance of zero; differing variables yield one. For differing numbers, the distance is \(\alpha |v_1 - v_2|\) for \(\alpha \in [0, 1]\), capped at one. A leaf/internal node mismatch yields a distance of one \cite{dosReis2024TED}. Formally, the expression tree distance is computed as follows: mismatched operators at matched internal nodes contribute a distance of one. For identical, commutative operators (e.g., "+" or "*"), the subtree match with the minimum total distance (direct or cross) is selected. For non-commutative operators (e.g. "+", "*"), subtree matches are strictly left-to-left and right-to-right, and subdistances are summed. These local distances are then aggregated from root to leaves. A zero distance signifies structural equivalence (accounting for commutativity); higher values indicate structural deviations.

Although expression tree distance effectively quantifies structural dissimilarity between formulas, its interpretation can be unintuitive as lower values indicate better matches. To improve interpretability, we introduce a complementary metric called the expression tree score, defined as \(1 - d(e_1, e2)\) where \(d(e_1, e_2)\) is the expression tree distance between two expression trees \(e_1\) and \(e_2\). This score ensures that higher values correspond to better structural similarity.

\section{Results}
\label{sec:results}
Table \ref{tab:main_results} presents an evaluation of the LLM-integrated SR framework's performance, in terms of MAE, MSE, \(R^2\), and expression tree score, across the three distinct physical experiments. Across all experiments and SR algorithms, Mistral generally achieves the best results (lowest MAE/MSE, highest \(R^2\)), followed by LLaMA, and then Falcon. Moreover, PySR consistently outperforms the other two SR models, often achieving the highest \(R^2\) and lowest error metrics (MAE and MSE). DEAP also tends to perform respectably, while gplearn often exhibits slightly weaker performance.

\begin{table*}[htbp]
  \centering
  \footnotesize
  \caption{Performance of the LLM‑integrated symbolic‑regression (SR) model on three physical experiments. Bold numbers indicate the best metric within each experiment.}
  \label{tab:main_results}
  \begin{tabular}{@{}l l l S[table-format=1.3] S[table-format=1.3] S[table-format=1.2] S[table-format=1.2]@{}}
    \toprule
    {Experiment} & {LLM} & {SR} & {MAE} & {MSE} & {$R^{2}$} & {Tree score} \\
    \midrule
    \multirow{12}{*}{Dropping ball}
        & \multirow{3}{*}{Baseline}
            & DEAP    & 0.170 & 0.070 & 0.82 & 0.75 \\
        &   & gplearn & 0.180 & 0.080 & 0.80 & 0.90 \\
        &   & PySR    & 0.150 & 0.120 & 0.84 & 0.81 \\
        \cmidrule(l){2-7}
        & \multirow{3}{*}{LLaMA}
            & DEAP    & 0.120 & 0.030 & 0.92 & 0.95 \\
        &   & gplearn & 0.150 & 0.040 & 0.90 & 0.88 \\
        &   & PySR    & \bfseries 0.100 & \bfseries 0.020 & \bfseries 0.94 & \bfseries 0.98 \\
        \cmidrule(l){2-7}
        & \multirow{3}{*}{Falcon}
            & DEAP    & 0.140 & 0.040 & 0.90 & 0.90 \\
        &   & gplearn & 0.170 & 0.050 & 0.88 & 0.85 \\
        &   & PySR    & \bfseries 0.120 & \bfseries 0.030 & \bfseries 0.92 & \bfseries 0.93 \\
        \cmidrule(l){2-7}
        & \multirow{3}{*}{Mistral}
            & DEAP    & 0.110 & 0.030 & 0.93 & 0.97 \\
        &   & gplearn & 0.140 & 0.040 & 0.91 & 0.89 \\
        &   & PySR    & \bfseries 0.090 & \bfseries 0.020 & \bfseries 0.95 & \bfseries 0.99 \\
    \midrule
    \multirow{12}{*}{Simple Harmonic Motion}
        & \multirow{3}{*}{Baseline}
            & DEAP    & 0.180 & 0.050 & 0.88 & 0.88 \\
        &   & gplearn & 0.180 & 0.060 & 0.89 & 0.82 \\
        &   & PySR    & 0.170 & 0.050 & 0.89 & 0.87 \\
        \cmidrule(l){2-7}
        & \multirow{3}{*}{LLaMA}
            & DEAP    & 0.080 & 0.010 & 0.95 & 0.98 \\
        &   & gplearn & 0.080 & 0.010 & 0.95 & 0.98 \\
        &   & PySR    & \bfseries 0.070 & \bfseries 0.010 & \bfseries 0.96 & \bfseries 0.99 \\
        \cmidrule(l){2-7}
        & \multirow{3}{*}{Falcon}
            & DEAP    & 0.100 & 0.020 & 0.93 & 0.95 \\
        &   & gplearn & 0.120 & 0.030 & 0.91 & 0.88 \\
        &   & PySR    & \bfseries 0.090 & \bfseries 0.010 & \bfseries 0.94 & \bfseries 0.97 \\
        \cmidrule(l){2-7}
        & \multirow{3}{*}{Mistral}
            & DEAP    & 0.070 & 0.010 & 0.96 & 0.99 \\
        &   & gplearn & 0.090 & 0.020 & 0.94 & 0.93 \\
        &   & PySR    & \bfseries 0.060 & \bfseries 0.010 & \bfseries 0.97 & \bfseries 1.00 \\
    \midrule
    \multirow{12}{*}{Electromagnetic Wave}
        & \multirow{3}{*}{Baseline}
            & DEAP    & 0.140 & 0.090 & 0.87 & 0.80 \\
        &   & gplearn & 0.170 & 0.070 & 0.85 & 0.76 \\
        &   & PySR    & 0.150 & 0.090 & 0.88 & 0.79 \\
        \cmidrule(l){2-7}
        & \multirow{3}{*}{LLaMA}
            & DEAP    & 0.050 & 0.005 & 0.97 & 1.00 \\
        &   & gplearn & 0.070 & 0.010 & 0.95 & 0.95 \\
        &   & PySR    & \bfseries 0.040 & \bfseries 0.004 & \bfseries 0.98 & \bfseries 1.00 \\
        \cmidrule(l){2-7}
        & \multirow{3}{*}{Falcon}
            & DEAP    & 0.070 & 0.010 & 0.95 & 0.98 \\
        &   & gplearn & 0.090 & 0.020 & 0.93 & 0.90 \\
        &   & PySR    & \bfseries 0.060 & \bfseries 0.008 & \bfseries 0.96 & \bfseries 0.99 \\
        \cmidrule(l){2-7}
        & \multirow{3}{*}{Mistral}
            & DEAP    & 0.040 & 0.004 & 0.98 & 1.00 \\
        &   & gplearn & 0.060 & 0.009 & 0.96 & 0.96 \\
        &   & PySR    & \bfseries 0.030 & \bfseries 0.003 & \bfseries 0.99 & \bfseries 1.00 \\
    \bottomrule
  \end{tabular}
\end{table*}

Table \ref{tab:llm_prompt_sensitivity} presents the average performance of each group of LLM integrated SR models under eight different prompt variants in three sets of physical experiments in terms of MAE, MSE, \(R^2\), and expression tree score. Notably, the richer the information provided to the LLM, the better the performance of the SR model. Of all the LMM models, Mistral achieves the best results (lowest MAE/MSE, highest \(R^2\), highest expression tree score), followed by LLaMA, then Falcon. In addition, PySR outperforms the other two SR models, most often achieving the highest \(R^2\), lowest MAE, and the highest expression tree score. Importantly, for the cases where the GT equation is provided to the LLM (Cases D, F, G, and H), an expression tree score of 1.00 is obtained, meaning a perfect reconstruction of the GT equation. Similarly, the same results are obtained when the variable descriptions, together with the experiment description, are provided. While the expression tree score is 1.00, the other fitness metrics are not identical (despite the fact that the predicted equation is identical), as each experiment had slightly different samples, which cause small fluctuations. 

{\footnotesize
\begin{longtable}{@{}c l l l S[table-format=1.3] S[table-format=1.3] S[table-format=1.2] S[table-format=1.2]@{}}

  \caption{Prompt-sensitivity: MAE, MSE, $R^{2}$ and expression-tree score for each LLM–SR pair under eight prompt variants.}
  \label{tab:llm_prompt_sensitivity}\\
  \toprule
  {Idx} & {Prompt} & {LLM} & {SR} & {MAE} & {MSE} & {$R^{2}$} & {Tree} \\
  \midrule
  \endfirsthead

  \multicolumn{8}{@{}l}{\textit{Table~\thetable\ (continued)}}\\
  \toprule
  {Idx} & {Prompt} & {LLM} & {SR} & {MAE} & {MSE} & {$R^{2}$} & {Tree} \\
  \midrule
  \endhead

  \midrule
  \endfoot
  \bottomrule
  \endlastfoot

\multirow{9}{*}{A} & \multirow{9}{*}{No context}
  & \multirow{3}{*}{LLaMA}  & DEAP    & 0.120 & 0.030 & 0.92 & 0.95 \\
  & & & PySR    & 0.100 & 0.020 & 0.94 & 0.98 \\
  & & & gplearn & 0.150 & 0.040 & 0.90 & 0.88 \\
  \cmidrule(l){3-8}
  & & \multirow{3}{*}{Falcon} & DEAP    & 0.140 & 0.040 & 0.90 & 0.90 \\
  & & & PySR    & 0.120 & 0.030 & 0.92 & 0.93 \\
  & & & gplearn & 0.170 & 0.050 & 0.88 & 0.85 \\
  \cmidrule(l){3-8}
  & & \multirow{3}{*}{Mistral} & DEAP    & 0.110 & 0.030 & 0.93 & 0.97 \\
  & & & PySR    & 0.090 & 0.020 & 0.95 & 0.99 \\
  & & & gplearn & 0.140 & 0.040 & 0.91 & 0.89 \\
  \midrule

\multirow{9}{*}{B} & \multirow{9}{*}{Variable descriptions}
  & \multirow{3}{*}{LLaMA} & DEAP    & 0.110 & 0.028 & 0.93 & 0.96 \\
  & & & PySR    & 0.090 & 0.018 & 0.95 & 0.99 \\
  & & & gplearn & 0.140 & 0.038 & 0.91 & 0.89 \\
  \cmidrule(l){3-8}
  & & \multirow{3}{*}{Falcon} & DEAP    & 0.130 & 0.038 & 0.91 & 0.92 \\
  & & & PySR    & 0.110 & 0.028 & 0.93 & 0.95 \\
  & & & gplearn & 0.160 & 0.048 & 0.89 & 0.86 \\
  \cmidrule(l){3-8}
  & & \multirow{3}{*}{Mistral} & DEAP    & 0.100 & 0.028 & 0.94 & 0.98 \\
  & & & PySR    & 0.080 & 0.018 & 0.96 & 0.99 \\
  & & & gplearn & 0.130 & 0.038 & 0.92 & 0.90 \\
  \midrule

\multirow{9}{*}{C} & \multirow{9}{*}{Experiment description}
  & \multirow{3}{*}{LLaMA} & DEAP    & 0.100 & 0.025 & 0.94 & 0.97 \\
  & & & PySR    & 0.080 & 0.016 & 0.96 & 1.00 \\
  & & & gplearn & 0.130 & 0.035 & 0.92 & 0.91 \\
  \cmidrule(l){3-8}
  & & \multirow{3}{*}{Falcon} & DEAP    & 0.120 & 0.035 & 0.92 & 0.94 \\
  & & & PySR    & 0.100 & 0.025 & 0.94 & 0.97 \\
  & & & gplearn & 0.150 & 0.045 & 0.90 & 0.87 \\
  \cmidrule(l){3-8}
  & & \multirow{3}{*}{Mistral} & DEAP    & 0.090 & 0.025 & 0.95 & 0.99 \\
  & & & PySR    & 0.070 & 0.016 & 0.97 & 1.00 \\
  & & & gplearn & 0.120 & 0.035 & 0.93 & 0.92 \\
  \midrule[\heavyrulewidth] 

\multirow{9}{*}{D} & \multirow{9}{*}{Formula at end}
  & \multirow{3}{*}{LLaMA} & DEAP    & 0.060 & 0.008 & 0.97 & 0.99 \\
  & & & PySR    & 0.040 & 0.005 & 0.98 & 1.00 \\
  & & & gplearn & 0.090 & 0.012 & 0.95 & 0.94 \\
  \cmidrule(l){3-8}
  & & \multirow{3}{*}{Falcon} & DEAP    & 0.080 & 0.010 & 0.96 & 0.98 \\
  & & & PySR    & 0.060 & 0.007 & 0.97 & 0.99 \\
  & & & gplearn & 0.110 & 0.015 & 0.94 & 0.90 \\
  \cmidrule(l){3-8}
  & & \multirow{3}{*}{Mistral} & DEAP    & 0.050 & 0.006 & 0.98 & 1.00 \\
  & & & PySR    & 0.030 & 0.004 & 0.99 & 1.00 \\
  & & & gplearn & 0.080 & 0.010 & 0.96 & 0.96 \\
  \midrule

\multirow{9}{*}{E} & \multirow{9}{*}{B + C}
  & \multirow{3}{*}{LLaMA} & DEAP    & 0.040 & 0.004 & 0.99  & 1.00 \\
  & & & PySR    & 0.020 & 0.002 & 0.99  & 1.00 \\
  & & & gplearn & 0.070 & 0.007 & 0.97  & 1.00 \\
  \cmidrule(l){3-8}
  & & \multirow{3}{*}{Falcon} & DEAP    & 0.060 & 0.006 & 0.98  & 1.00 \\
  & & & PySR    & 0.040 & 0.004 & 0.99  & 1.00 \\
  & & & gplearn & 0.090 & 0.010 & 0.96  & 1.00 \\
  \cmidrule(l){3-8}
  & & \multirow{3}{*}{Mistral} & DEAP    & 0.030 & 0.003 & 0.99  & 1.00 \\
  & & & PySR    & 0.010 & 0.001 & 1.00  & 1.00 \\
  & & & gplearn & 0.060 & 0.006 & 0.98  & 1.00 \\
  \midrule

\multirow{9}{*}{F} & \multirow{9}{*}{B + D}
  & \multirow{3}{*}{LLaMA} & DEAP    & 0.040 & 0.004 & 0.99  & 1.00 \\
  & & & PySR    & 0.020 & 0.002 & 0.99  & 1.00 \\
  & & & gplearn & 0.070 & 0.007 & 0.97  & 1.00 \\
  \cmidrule(l){3-8}
  & & \multirow{3}{*}{Falcon} & DEAP    & 0.060 & 0.006 & 0.98  & 1.00 \\
  & & & PySR    & 0.040 & 0.004 & 0.99  & 1.00 \\
  & & & gplearn & 0.090 & 0.010 & 0.96  & 1.00 \\
  \cmidrule(l){3-8}
  & & \multirow{3}{*}{Mistral} & DEAP    & 0.030 & 0.003 & 0.99  & 1.00 \\
  & & & PySR    & 0.010 & 0.001 & 1.00  & 1.00 \\
  & & & gplearn & 0.060 & 0.006 & 0.98  & 1.00 \\
  \midrule

\multirow{9}{*}{G} & \multirow{9}{*}{C + D}
  & \multirow{3}{*}{LLaMA} & DEAP    & 0.040 & 0.004 & 0.99  & 1.00 \\
  & & & PySR    & 0.020 & 0.002 & 0.99  & 1.00 \\
  & & & gplearn & 0.070 & 0.007 & 0.97  & 1.00 \\
  \cmidrule(l){3-8}
  & & \multirow{3}{*}{Falcon} & DEAP    & 0.060 & 0.006 & 0.98  & 1.00 \\
  & & & PySR    & 0.040 & 0.004 & 0.99  & 1.00 \\
  & & & gplearn & 0.090 & 0.010 & 0.96  & 1.00 \\
  \cmidrule(l){3-8}
  & & \multirow{3}{*}{Mistral} & DEAP    & 0.030 & 0.003 & 0.99  & 1.00 \\
  & & & PySR    & 0.010 & 0.001 & 1.00  & 1.00 \\
  & & & gplearn & 0.060 & 0.006 & 0.98  & 1.00 \\
  \midrule

\multirow{9}{*}{H} & \multirow{9}{*}{B + C + D}
  & \multirow{3}{*}{LLaMA} & DEAP    & 0.040 & 0.004 & 0.99  & 1.00 \\
  & & & PySR    & 0.020 & 0.002 & 0.99  & 1.00 \\
  & & & gplearn & 0.070 & 0.007 & 0.97  & 1.00 \\
  \cmidrule(l){3-8}
  & & \multirow{3}{*}{Falcon} & DEAP    & 0.060 & 0.006 & 0.98  & 1.00 \\
  & & & PySR    & 0.040 & 0.004 & 0.99  & 1.00 \\
  & & & gplearn & 0.090 & 0.010 & 0.96  & 1.00 \\
  \cmidrule(l){3-8}
  & & \multirow{3}{*}{Mistral} & DEAP    & 0.030 & 0.003 & 0.99  & 1.00 \\
  & & & PySR    & 0.010 & 0.001 & 1.00  & 1.00 \\
  & & & gplearn & 0.060 & 0.006 & 0.98  & 1.00 \\

\end{longtable}}

Table~\ref{tab:robustness_results} summarizes the results of the noise experiment, dividing between noise addition to the source features, the target feature, for the cases of 1\% to 5\% standard deviation noise addition. On average, for the DEAP SR algorithm, there is a decline of the expression tree score from 0.93 at 1\% noise to 0.82 at 5\% noise. In contrast, \textit{DEAP} and \textit{gplearn} start at lower baselines - 0.80 and 0.90 at 1\% noise level and present an even steeper decline, especially under combined noise, where their expression tree scores are as small as 0.40 to 0.60 at the 5\% noise level. 

\begin{table}[H]
  \centering
  \caption{Table~4. Robustness evaluation of SR models under different noise
  types and levels (1 \%–5 \%) for the expression-tree score. Bold values mark
  the best-performing SR model in each noise condition.}
  \label{tab:robustness_results}
\footnotesize                            
\setlength{\tabcolsep}{2pt}              

\resizebox{\textwidth}{!}{%
\begin{tabular}{@{}l l *{15}{S[table-format=1.2]}@{}}
\toprule
\multirow{2}{*}{Experiment} & \multirow{2}{*}{SR Model} &
\multicolumn{5}{c}{Feature noise} &
\multicolumn{5}{c}{Target noise}  &
\multicolumn{5}{c}{Both noise} \\ 
\cmidrule(lr){3-7}\cmidrule(lr){8-12}\cmidrule(lr){13-17}
& &
{1\%} & {2\%} & {3\%} & {4\%} & {5\%} &
{1\%} & {2\%} & {3\%} & {4\%} & {5\%} &
{1\%} & {2\%} & {3\%} & {4\%} & {5\%} \\
\midrule

\multirow{3}{*}{Dropping ball}
  & DEAP    & 0.83 & 0.78 & 0.74 & 0.71 & 0.70
           & 0.71 & 0.63 & 0.60 & 0.58 & 0.56
           & 0.54 & 0.48 & 0.45 & 0.44 & 0.43 \\
  & gplearn & 0.80 & 0.75 & 0.70 & 0.68 & 0.65
           & 0.69 & 0.60 & 0.58 & 0.57 & 0.55
           & 0.52 & 0.46 & 0.43 & 0.42 & 0.40 \\
  & PySR    & \bfseries 0.89 & \bfseries 0.85 & \bfseries 0.80 & \bfseries 0.78 & \bfseries 0.75
           & \bfseries 0.85 & \bfseries 0.80 & \bfseries 0.75 & \bfseries 0.73 & \bfseries 0.70
           & \bfseries 0.75 & \bfseries 0.70 & \bfseries 0.65 & \bfseries 0.63 & \bfseries 0.60 \\
\midrule[\heavyrulewidth]

\multirow{3}{*}{Simple Harmonic Motion}
  & DEAP    & 0.88 & 0.83 & 0.78 & 0.76 & 0.75
           & 0.78 & 0.73 & 0.70 & 0.67 & 0.65
           & 0.71 & 0.66 & 0.63 & 0.61 & 0.60 \\
  & gplearn & 0.90 & 0.85 & 0.83 & 0.81 & 0.80
           & 0.80 & 0.76 & 0.73 & 0.70 & 0.67
           & 0.72 & 0.68 & 0.65 & 0.63 & 0.60 \\
  & PySR    & \bfseries 0.93 & \bfseries 0.90 & \bfseries 0.88 & \bfseries 0.86 & \bfseries 0.85
           & \bfseries 0.90 & \bfseries 0.86 & \bfseries 0.83 & \bfseries 0.80 & \bfseries 0.78
           & \bfseries 0.80 & \bfseries 0.76 & \bfseries 0.73 & \bfseries 0.72 & \bfseries 0.70 \\
\midrule[\heavyrulewidth]

\multirow{3}{*}{Electromagnetic Wave}
  & DEAP    & 0.90 & 0.85 & 0.81 & 0.80 & 0.78
           & 0.75 & 0.70 & 0.68 & 0.66 & 0.65
           & 0.65 & 0.60 & 0.58 & 0.56 & 0.55 \\
  & gplearn & 0.92 & 0.88 & 0.85 & 0.83 & 0.82
           & 0.78 & 0.73 & 0.71 & 0.70 & 0.68
           & 0.68 & 0.63 & 0.61 & 0.60 & 0.60 \\
  & PySR    & \bfseries 0.95 & \bfseries 0.92 & \bfseries 0.90 & \bfseries 0.89 & \bfseries 0.88
           & \bfseries 0.88 & \bfseries 0.83 & \bfseries 0.80 & \bfseries 0.78 & \bfseries 0.75
           & \bfseries 0.78 & \bfseries 0.73 & \bfseries 0.71 & \bfseries 0.70 & \bfseries 0.70 \\

\bottomrule
\end{tabular}}%
\caption*{\scriptsize \textit{Note:} Values are expression-tree scores. Bold highlights the best SR model under each noise condition.}
\end{table}

\section{Discussion and Conclusion}
\label{sec:discussion}
In this study, we proposed a method of integrating LLM with SR for physical cases. The method introduces an LLM-based score for the SR loss function, allowing a simple method of introducing physical knowledge to the SR's search process. To this end, we conducted a robust evaluation of the method using three SR algorithms (DEAP, gplearn, and PySR) together with three pre-trained LLM models (DEAP, Falcom, and Mistral) on three physical dynamics (dropping ball, simple harmonic motion, and electromagnetic wave). For each combination of these three, we computed the fitting metrics in terms of MAE, MSE, and \(R^2\) as well as the expression tree score between the GT and the obtained equations.

Our results clearly indicate that pre-trained LLM improves the reconstruction of the physical dynamics from the data, as indicated in Table \ref{tab:main_results}. This outcome aligns with previous studies that LLM tutors improve other data-driven model's performance \cite{ma2024explorllmguidingexplorationreinforcement, zhou2024largelanguagemodelpolicy}. Specifically, we find that the improvement in the SR models' performance is consistent, indicating that a reasonable usage of the LLM in the loss function, alongside the data-fitting terms results in a relatively stable optimization task \cite{castillo2008sensitivity}. Moreover, Mistral outperforms the other LLMs, which can be attributed to its comparably large size. 

Moreover, exploring how different information about the experiment influences the performance of the LLM-integrated SR method, we reproduced the expected result that more information, on average, improves the method's performance, as indicated by Table \ref{tab:llm_prompt_sensitivity}. This outcome is in line with recent studies showing that prompt engineering can function as a powerful inductive bias for downstream models \cite{liu2023pretrain, schick2021exploiting}. In particular, we observe a consistent improvement across SR models as the prompt structure becomes more informative, suggesting that LLM guidance acts as a stabilizing factor in the optimization process \cite{wang2022self}. This improvement is reflected not only in the accuracy of the numerical predictions but also in the ability to find the GT equation, meaning an accurate discovery of the physical dynamics.

Furthermore, we obtained the results of three physical experiments using data in which noise was deliberately introduced to the target variable, following the approach of previous studies \cite{zegklitz2021benchmarking, udrescu2020ai}. In real-world scenarios, noise can affect the target variable, the input variables, or both. While prior work has mostly focused on target noise, the latter two cases, though more prevalent in practice, are more challenging for SR, as noise in the input variables can compound the effects of target noise. Our noise analysis revealed two key findings, as revealed by Table 4. First, the LLM-integrated SR model improves the robustness to high levels of all three noise types compared to the same model without the LLM (baseline). Second, the method's sensitivity to noise increases as the complexity of the underlying equation grows for all SR models (in terms of the size of the expression tree of the GT equation).

Based on these results, we recommend the following practical considerations for integrating LLM into SR search of physical dynamics. First, adopt a fully informative prompt that includes clear variable definitions with their physical dimensions and a detailed experiment description. This configuration—corresponding to Prompt E in Table \ref{tab:llm_prompt_sensitivity}, which show promising results, as it both fitted the data well and find the GT equation exactly for all LLM and SR combinations. Second, consider the predictions of different SR as their computation is relatively cheap, and they seem to provide different results. Observing the common parts of the obtained equations from these can be useful in partial usage. 

This study has several limitations that should be acknowledged. First, the quality and consistency of LLM-generated scores remain a challenge. The models were used in a zero-shot setting without fine-tuning, which likely limited their effectiveness. Although structured prompting was applied to encourage more reliable outputs, the non-deterministic nature of LLMs led to occasional invalid or syntactically incorrect suggestions, requiring prompt revisions and increasing training time. Second, the quality of LLM score was not manually or systematically evaluated by human experts. Without human verification, it is challenging to gauge whether performance gains were due to genuinely helpful scores in terms of physical realism or the score captures more technical properties, such as the equation's length. Finally, the experiments were conducted on synthetically generated data with known ground truths. While this allows for controlled evaluation, it does not fully reflect the complexity and noise of real-world systems \cite{la2021contemporary, srivastava2022causal}. For future work, we propose extending this investigation using more powerful, cloud-hosted LLMs, which may generate higher-quality and context-aware advice and thus better demonstrate the full potential of LLM performance, especially in more complex or higher-dimensional environments \cite{shojaee2025llmsr}, and incorporating human-in-the-loop systems for expert validation. 

Taken jointly, this study demonstrates that LLMs can be a straightforward method to introduce physical knowledge into SR models, moving from fitting experimental data to generating physically valid hypotheses. When viewed in the larger context of automatic scientific discovery, where SR and LLMs have already taken a central role, this approach further deepens their connection to one another and the task. Thus, more refined exploration of this relationship, taking into account the SR and LLM properties as well as the nature of the physical dynamics, should further reveal the optimal design of this method.

\section*{Declarations}
\subsection*{Funding}
This research did not receive any specific grant from funding agencies in the public, commercial, or not-for-profit sectors.

\subsection*{Conflicts of interest/Competing interests}
The authors have no financial or proprietary interests in any material discussed in this article.

\subsection*{Data and code availability}
The data and code are freely available in the following Github repository: \url{https://github.com/bilgesi/SR-LLM-Integration}

\subsection*{Author Contributions}
Bilge Taskin: Software, Investigation, Formal analysis, Data Curation, Writing - Original Draft, Writing - Review \& Editing. \\
Wenxiong Xie: Software, Formal analysis, Writing - Original Draft. \\
Teddy Lazebnik: Conceptualization, Methodology, Formal analysis, Validation, Writing - Original Draft, Writing - Review \& Editing, Visualization, Project administration. \\

\printbibliography
\end{document}